# Identification and classification of TCM syndrome types among patients with vascular mild cognitive impairment using latent tree analysis

Chen Fu, M.D.: Dongfang Hospital, Beijing University of Traditional Chinese Medicine; E-mail: fuchen2003@163.com

Nevin Lianwen Zhang, Ph.D.: The Hong Kong University of Science and Technology; E-mail:lzhang@cse.ust.hk

Bao Xin Chen, M.D.: Dongfang Hospital, Beijing University of Traditional Chinese Medicine; E-mail: baoxin.chen@gmail.com

Zhou Rong Chen, B.S.: The Hong Kong University of Science and Technology; E-mail: zchenbb@cse.ust.hk

Xiang Lan Jin, M.D.: Dongfang Hospital, Beijing University of Traditional Chinese Medicine; E-mail: jxlan2001@126.com

Rong Juan Guo, M.D.: Dongfang Hospital, Beijing University of Traditional Chinese Medicine; E-mail:dfguorongjuan@163.com

Zhi Gang Chen, M.D.: Dongfang Hospital, Beijing University of Traditional Chinese Medicine; E-mail: chenzhigang64@126.com

Yun Ling Zhang, M.D.: Dongfang Hospital, Beijing University of Traditional Chinese Medicine; E-mail: yunlingzhang2004@163.com

Corresponding Authors: Yun Ling Zhang, M.D.: Dongfang Hospital, Beijing University of Traditional Chinese Medicine; E-mail: yunlingzhang2004@163.com; Nevin L. Zhang, Department of Computer Science and Engineering, The Hong Kong University of Science and Technology; E-mail:lzhang@cse.ust.hk.

**Abstract**

*Objective:* To treat patients with vascular mild cognitive impairment (VMCI) using TCM, it is necessary to classify the patients into TCM syndrome types and to apply different treatments to different types. We investigate how to properly carry out the classification using a novel data-driven method known as latent tree analysis.

*Method:* A cross-sectional survey on VMCI was carried out in several regions in northern China from 2008 to 2011, which resulted in a data set that involves 803 patients and 93 symptoms. Latent tree analysis was performed on the data to reveal symptom co-occurrence patterns, and the patients were partitioned into clusters in multiple ways based on the patterns. The patient clusters were matched up with syndrome types, and population statistics of the clusters are used to quantify the syndrome types and to establish classification rules.

*Results:* Eight syndrome types are identified: Qi Deficiency, Qi Stagnation, Blood Deficiency, Blood Stasis, Phlegm-Dampness, Fire-Heat, Yang Deficiency, and Yin Deficiency. The prevalence and symptom occurrence characteristics of each syndrome type are determined. Quantitative classification rules are established for determining whether a patient belongs to each of the syndrome types.

*Conclusions:* A solution for the TCM syndrome classification problem associated with VMCI is established based on the latent tree analysis of unlabeled symptom survey data. The results can be used as a reference in clinic practice to improve the quality of syndrome differentiation and to reduce diagnosis variances across physicians. They can also be used for patient selection in research projects aimed at finding biomarkers for the syndrome types and in randomized control trials aimed at determining the efficacy of TCM treatments of VMCI.

1. **Introduction**

With the global increase in population and life expectancy, dementia has become a prominent world health problem. It is one of the main diseases that have significant impacts on the health and quality of life among seniors. In 2005, the global prevalence of dementia was estimated to be 32.85 million, and it is expected to double every 20 years, to 65.7 million in 2035 and 131.4 million in 2055 [1].

Dementia can be classified as degenerative, vascular, or other types based on its etiology. Up to 34% of dementia cases show significant vascular pathology [2-7]. Recently, the construct of "vascular cognitive impairment" (VCI) has been introduced to capture the entire spectrum of cognitive disorders, ranging from mild cognitive impairment to fully developed dementia, that are caused by or associated with vascular factors [8]. VCI has three subtypes: Vascular dementia (VaD), Alzheimer's disease (AD) with a vascular component (mixed dementia), and the pre-dementia stage that is referred to as vascular mild cognitive impairment (VMCI or VaMCI) [9].

There are currently no definitive checklists of diagnostic criteria for VCI and VMCI, and hence their prevalence are difficult to ascertain. However, it is known that the prevalence of VaD doubles every 5.3 years [10-11]. As the pre-dementia stage, VMCI is believed to be much more common.

Some studies have shown that 46% of patients with VMCI developed dementia after five years [12]. It is therefore important to apply some interventions to prevent and postpone VMCI in order to reduce the occurrence of dementia. The current treatment options for VMCI target at improving symptoms (e.g. memory loss, psychological symptoms such as depression and anxiety), or controlling the various risk factors in order to prevent cognitive function impairment. There are no effective medicines for the management of VMCI in Western medicine.

In traditional Chinese medicine (TCM), cognitive impairment is categorized as "dementia" (*chi dai* or *dai bing*). Acupuncture and other TCM treatments have been used to treat VMCI in China for years. Reports on a number of studies indicate that acupuncture techniques may be effective for improving intelligence, stimulating consciousness and enhancing memory, and appear to be beneficial for the management of cognitive and memory functions [13-15]. Findings from other studies suggest that oral Chinese Traditional Patent Medicine and comprehensive TCM intervention appear to improve the cognitive functions and quality of life for patients with cognitive impairment caused by leukoaraiosis [16-17].

To treat patients with VMCI using TCM, it is necessary to classify the patients into TCM syndrome types and prescribe different treatments for different types. The efficacy of TCM treatment depends heavily on whether the classification is done properly, an important issue that has so far received little attention [18]. In this paper, we seek to identify the TCM syndrome types among patients with VMCI and to establish classification rules for them using a novel method called latent tree analysis (LTA) [19].

The LTA method is a generalization of latent class analysis (LCA), a method often used in WM research to identify subtypes of a patient population in the absence of a gold standard [20,21]. LTA improves LCA by relaxing a key independence assumption made by LCA that is unrealistically strong.

2. **Data Collection**

Data used in this study were collected in a project on the TCM prevention and treatment of VMCI supported by the TCM Special Research Projects Program, China Ministry of Science and Technology under Contract No.200807011. Ethical approval for the project was given by The Ethics Committee on Clinical Research of Dongfang Hospital.

The project involved seven top-graded hospitals in northern China: The Dongfang Hospital of Beijing University of Chinese Medicine, Peking University People's Hospital, China Academy of Chinese Medical Sciences Affiliated Wangjing Hospital, Tianjin Medical University Second Affiliated Hospital, Shandong University of Traditional Chinese Medicine Affiliated Hospital, Changchun University of Chinese Medicine First Affiliated Hospital, and Hebei Medical University Traditional Chinese Medicine Hospital.

All VMCI patients in the project were recruited between February 2008 and February 2012 from the neurology divisions and wards of the hospitals, and the surrounding communities. Since there are no standard diagnostic criteria for VMCI, the research team set up the following criteria themselves based on the NINDS-CSN (National Institute of Neurological Disorders and Stroke–Canadian Stroke Network) vascular cognitive impairment harmonization standards [8]:

1) Age $\geq$ 50 years;

2) Presence of vascular factors (cerebral hemorrhage, cerebral infarction, leukoaraiosis, hypertension, diabetes, hyperlipoidemia, coronary heart disease, etc.);

3) Subjective complaints of cognitive impairment that were corroborated by an informant;

4) Objective impairment in more than one domains of cognitive abilities by neuropsychologic assessments, including MMSE, Clinical Dementia Rating (CDR), Montreal Cognitive Assessment (MoCA) Beijing version. The thresholds for these tests were set as follows: MMSE score > 17 points (illiterates), or > 20 points (education years $\leq$ 6 years), or > 24 points (education years > 6 years);  MoCA score < 25 points (education $\leq$ 12 years), or < 26 points (education > 12 years);  CDR score＝0 or 0.5 points;

5) Sustained cognitive decline for more than 6 months based on patient self-report or caregiver report;

6) Normal abilities or mild impairment of daily living activities/social activities according to the result of Activities of Daily Living Scale (ADL); and

7) Not meeting the criteria for the diagnosis of dementia in Statistical Manual of

Mental Disorders (DSM-IV-R).

The exclusion criteria included the following: Patients with mental illness or serious physical illness which could affect neuropsychological examinations; those with Hamilton Depression Rating Scale (HAMD) score $\geq$ 17 points; and those refused to participate in this study and sign on the informed-consent form.

The data collection teams consisted of neurology physicians and graduate students. The team members were trained in multiple sessions on various aspects of the data collection process and examined on the training materials. Only those who passed the examination were allowed to join the teams. The same training materials were used in all data collection centers.

Data collection was carried out by interviewing all patients using an interviewer administered questionnaire. The interviewers first decided whether a patient should be included in the study according to the inclusion and exclusion criteria listed above. They then moved on to collect information on TCM signs and symptoms and filled out the questionnaire. A total number of 803 patients were successfully interviewed and their demographic characteristics are summarized in Table 1.

The items in the questionnaire were extracted from both ancient and modern TCM literature on *chi dai* or *dai bing* patients. All signs and symptoms mentioned in the literature were initially included. The list was refined through multiple rounds of pilot surveys and expert consultations. Most items were polytomous and they were converted into dichotomous items for this analysis for simplicity, just as in the studies with LCA [20,21]. During preprocessing, a few items were combined. For example, edema on face and edema on limbs are combined into edema. The remaining 93 items are listed in Table 2. The items consist of signs and symptoms of interest to TCM. We will sometimes refer to them simply as symptoms for brevity. The data are unlabeled in the sense that the true syndrome types of the patients are unknown.

There was a quality control personnel at each data collection center to ensure the authenticity and completeness of the questionnaires. The project team paid regular visits to the data collection centers and randomly sampled questionnaires for authenticity, accuracy, and completeness checks. Data entry were carried out by two persons, each being responsible for entering half of the questionnaires and for verifying the other half.

3. **Method**

The VMCI data were studied using the LTA method described in [19]. The objective of the method is to reveal symptom patterns hidden in the data and to use those patterns to identify patient clusters that correspond to different TCM syndrome types. Statistical characteristics of the patient clusters are then used as quantitative characterizations of the syndrome types, and syndrome classification rules are thereby established.

The analysis was carried using the Lantern software [22]. There were five steps:

1. Pattern discovery: In the first step, latent tree analysis was performed to reveal symptom co-occurrence/mutual exclusion patterns hidden in the VMCI data. Lantern includes two algorithms for this purpose. In this study, the EAST (Extension Adjustment Simplification until Termination) algorithm [23] was used. The result is a latent tree model with multiple latent variables. Some of the latent variables reveal probabilistic symptom co-occurrence patterns, while others reveal probabilistic mutual-exclusion of symptoms.

2. Pattern Interpretation: The second step was to determine the TCM syndrome connotations of the patterns discovered. In the case of a symptom co-occurrence pattern, the task was to determine what TCM syndrome type(s) can bring about the co-occurrence of the symptoms in the pattern. The result could be a single syndrome type that can explain all the symptoms in the pattern, or a combination of several syndrome types, each explaining some of the symptoms. If a latent variable represents the probabilistic mutual exclusion of two or more groups of symptoms, each group is a co-occurrence pattern and was interpreted separately. Domain knowledge [24] was used in this step.

3. Syndrome Identification: The pattern interpretation process gave rise to a list of syndrome types. Each syndrome type explains a list of symptoms from various co-occurrence patterns. If all key manifestations of the syndrome type are present on the list, then it was regarded as well supported by the data and was selected as a target for further analysis. Domain knowledge was used in this step also.

4. Syndrome Quantification: For each syndrome type identified at the previous step, cluster analysis was performed on the patient population based on the symptoms that it explains. The symptom variables were not always used as features for the cluster analysis directly. If two or more symptom variables are from the same pattern, they were "combined" into one latent feature. The analysis divided the patients into several clusters. One patient cluster was identified as corresponding to the syndrome type. Population statistics of the patient cluster were used to quantify the syndrome type.

5. Syndrome Classification: Finally, a classification rule was established for each syndrome type. The rule can be used to determine whether a patient belongs to the syndrome type.

The main results，which will be presented in Sections 4.4 and 4.5，were obtained at the last two steps. The first three steps produced intermediate results that were necessary for the last two steps. They will be presented in Sections 4.1 to 4.3, along with discussions of some subtle practical issues that are not addressed in [19]. Those are important for researchers who wish to apply the method in their own work.

# 4 Results

## 4.1 Pattern Discovery

Latent tree analysis of the VMCI data yielded a model with 32 latent variables.

The structure of the model is shown in Figure 3 in [19]. Some of the latent variables reveal probabilistic symptom co-occurrence (i.e., positive correlation) patterns, while others reveal probabilistic mutual exclusion (i.e., negative correlation) of symptoms. The patterns were extracted from the model using the Model Interpretation function of Lantern, which is explained in [19].

The co-occurrence patterns discovered are shown in the second column of Table 3. We see, for example, that the latent variable Y29 reveals the probabilistic co-occurrence of the three symptoms lack of strength, mental fatigue, and loose stool; Y30 reveals the probabilistic co-occurrence of frequent nocturnal urination and dripping urination; Y31 reveals the probabilistic co-occurrence of blurred vision and dry eyes; and so on. Note that there is only one symptom for Y23 in the table. This is because Y23 is directly connected to only one symptom variable in the model.

Each co-occurrence pattern is given as a list of symptoms. Ordering in the list is important. Conceptually, a pattern corresponds to a class of patients [19]. Symptoms at the front of the list are more important than those at the end in determining whether a patient belongs to the cluster. For example, the list for the pattern Y27 consists of short of breath, sighing, hypochondrium distension or pain, tinnitus resemble tide, and stabbing headache. Those symptoms are ordered in descending importance, with short of breath being the most important and stabbing headache being the least important.

The mutual-exclusion patterns discovered are shown in the second column of Table 4. For example, Y02 reveals the probabilistic mutual exclusion of two groups of symptoms: Y02-1 (anorexia, undigested food in stool, etc.) and Y02-2 (darkish tongue). The first group Y02-1 consists of more than one symptom. They are positively correlated and hence tend to co-occur. In other words, Y02-1 is a co-occurrence pattern. Similarly, Y09-1 (slow pulse, moderate pulse) and Y25-1 (insomnia, dreamfulness) are also co-occurrence patterns.

Y14 and Y15 each have three states. They reveal the mutual exclusion of thirst desire no drinks, thirst desire cold drinks, and thirst desire hot drinks. The symptom dry mouth or throat is a higher level symptom. It is subsumed by the forgoing three symptoms and hence removed from further analysis.

### 4.2 Pattern Interpretation

To interpret the co-occurrence of a group of symptoms amounts to identifying the syndrome type(s) that can lead to the co-occurrence of the symptoms. Often there is no syndrome type that can explain the occurrence of all the symptoms in a pattern. Consequently, we have primary interpretations and secondary interpretations. A primary interpretation attempts to explain the co-occurrence of all symptoms in a pattern, and, when that is not possible, tries to explain the leading symptom and as many subsequent symptoms as possible. A secondary interpretation explains other symptoms or gives alternative explanations to some of the symptoms.

The last two columns of Tables 3 and 4 give interpretations of all the patterns discovered by latent tree analysis. For example, Y01 reveals the probabilistic co-occurrence of four symptoms: Asthenia of defecation, dry stool or constipation, sallow complexion, and clear profuse urination. No single syndrome types can explain all the four symptoms. The primary interpretation is Qi Deficiency and it explains the co-occurrence of the first two symptoms. Blood Deficiency is a secondary interpretation and it explains the co-occurrence of the second and the third symptoms. In addition, dry stool or constipation can also be caused by Fire-Heat, Yin Deficiency and Qi Stagnation. Clear profuse urination is weakly correlated with the other three symptoms and it is a manifestation of Yang Deficiency.

As the second example, consider Y06. It reveals the probabilistic co-occurrence of the two symptoms thick tongue fur and greasy tongue fur. The syndrome type Phlegm-Dampness explains both symptoms. It is the primary interpretation of the pattern. There are no secondary interpretations.

As the third example, consider Y25-1. It is about the probabilistic co-occurrence of insomnia and dreamfulness. The two symptoms can be explained by any of the following five syndrome types: Yin Deficiency, Fire-Heat, Blood Deficiency, Qi Deficiency, and Phlegm-Dampness. They are all primary and alternative interpretations of the pattern. There are no secondary interpretations.

The above examples represent three possible scenarios one might encounter when interpreting a pattern. First, the pattern has only one primary interpretation and no secondary interpretations. The patterns Y06, Y08, Y18, Y23 and Y29, and Y02-2, Y05-2, Y09-1, Y09-2, Y10-1,Y12-1, Y14-2 and Y15-1 fall into this category. Second, the pattern has more than one primary interpretations and no secondary interpretations. The patterns Y13, Y31, Y12-2, Y14-1, Y25-1 and Y25-2 belong to this category. Third, the pattern has both primary interpretations and secondary interpretations. The patterns Y01, Y04, Y03, Y07, Y11, Y16, Y17, Y19, Y20, Y21, Y22, Y24, Y26, Y27, Y30, Y02-1 and Y10-2 are in this category.

The interpretation of Y14-1 (thirst desire no drinks) requires some discussions. According to TCM theory, the symptom can be due to Blood Stasis. In the VMCI data set, however, it is found (in the syndrome quantification step) to be negatively correlated with other manifestations of Blood Stasis such as purples or darkish lips, pale complexion, and blackish eyelid. We therefore conclude that, for the population under study, thirst desire no drinks is caused by other factors, namely Yin Deficiency and Phlegm-Dampness, rather than Blood Stasis. Blood Stasis is hence not included in Table 4 as an interpretation for the symptom. Similarly, the pattern Y09-1 (slow pulse, moderate pulse) can be explained by either Qi Deficiency or Phlegm-Dampness in theory. However, it is negatively correlated with other manifestations of Phlegm-Dampness such as thick tongue and greasy tongue fur, and hence Phlegm-Dampness is not listed as a possible explanation for the pattern. The pattern Y10-1 (pink tongue, hollow headache) can be explained by Qi Deficiency, Yang Deficiency, or Yin Deficiency. However, it is negatively correlated with other manifestations of Qi Deficiency and Yang Deficiency, and hence only Yin Deficiency is listed as its primary interpretation.

In addition, several patterns are not interpreted. Y05-1 (white tongue fur) is not interpreted because it considered normal for the population under study, namely seniors aged 60 or above and with VMCI. Y15-2 is not interpreted because it is subsumed by Y14-1 and Y14-2.

**4.3 Syndrome Identification**

Based on the patterns and their interpretations in Tables 3 and 4, we have identified eight syndrome types that are well supported by the data: Qi Deficiency, Qi Stagnation, Blood Deficiency, Blood Stasis, Phlegm-Dampness, Fire-Heat, Yang Deficiency, and Yin Deficiency.

Take Yang Deficiency as an example. According to Tables 3 and 4, here are the symptom groups that it explains:

> Y24 (lassitude of limbs or body, fear of cold or cold limbs), Y26 (chest oppression, palpitation), Y30 (frequent nocturnal urination), Y02-1 (anorexia, undigested food in stool, bland taste in mouth, vomiting of saliva, diarrhea before dawn), Y15-1 (thirst desire hot drinks); Y01 (clear profuse urination), Y03 (pale complexion), Y04 (spontaneous sweating), Y07 (muscular twitching), Y19 (sore waist or knees), Y22 (dim complexion, blackish lower eyelid).

There is a semicolon in the list. The symptom groups before the semicolon are from patterns for which the syndrome type is a primary interpretation, whereas the symptom groups after the semicolon are from patterns for which the syndrome type is a secondary interpretation.

Those symptom groups cover different aspects of the impacts of Yang Deficiency, including its impacts on kidney (Y19), on the digestive system (Y02-1), on urination and drinking (Y30, Y15-1), on muscle (Y07), and on the striae and interstice (Y04), as well as its manifestations on the body and limbs (Y24), in the chest (Y26), and on the face (Y03, Y22). Since major aspects of the impacts of Yang Deficiency are covered, we conclude that there is sufficient information in the data to quantify Yang

Deficiency.

The above list is quite long and the symptoms are not in any particular order. In the next section, we will order the symptoms according to their importance in characterizing Yang Deficiency. In addition, many of the symptoms will be pruned using the criterion of cumulative information coverage [19]. The result will be a clean and quantitative characterization of Yang Deficiency. Although it is not part of the final result, making the list is a necessary intermediate step.

The lists of symptom groups that support each of the other seven syndrome types are listed below.

- Blood Deficiency: Y03 (pale lips, pale complexion, dizzy headache, thin tongue, hemihidrosis), Y31 (blurred vision, dry eyes), Y12-2 (thin pulse), Y25-1 (insomnia, dreamfulness); Y01 (dry stool or constipation, sallow complexion), Y07 (muscular twitching), Y17 (trembling of limbs), Y21 (dizziness), Y24 (numbness), Y26 (palpitation).

- Blood Stasis: Y13 (varicose sublingual-veins, astringent pulse, tense pulse), Y22 (dim complexion, blackish lower eyelid, scaly skin), Y23 (purples or darkish lips), Y02-2 (darkish tongue); Y20 (tongue with ecchymosis), Y21 (dizziness), Y24 (numbness), Y26 (palpitation), Y27 (stabbing headache).

- Qi Deficiency: Y01 (asthenia of defecation, dry stool or constipation), Y08 (sunken pulse, feeble pulse), Y20 (fat tongue, tooth-marked tongue), Y26 (chest oppression, palpitation), Y29 (lack of strength, mental fatigue, loose stool), Y09-1 (slow pulse, moderate pulse), Y12-2 (thin pulse), Y25-1 (insomnia, dreamfulness); Y02 (anorexia, bland taste in mouth), Y03 (pale complexion, dizzy headache, loose stool following dry feces), Y04 (spontaneous sweating), Y16 (urinary incontinence,) Y19 (sore waist or knees), Y21 (dizziness), Y24 (lassitude of body and limbs), Y27 (short of breath), Y30 (dripping urination).

- Qi Stagnation: Y13 (varicose sublingual-veins, astringent pulse, tense pulse),

Y27 (short of breath, sighing, hypochondrium distention or pain), Y09-2 (taut pulse); Y01 (dry stool or constipation), Y17 (abdominal distension), Y26 (chest oppression).

- Fire-Heat: Y16 (bitter taste in mouth, urinary incontinence), Y17 (acid swallow or epigastric upset, trembling of limbs), Y18 (aphtha on month or tongue, throbbing headache), Y21 (dizziness, distending headache, nausea or vomiting), Y05-2 (yellow tongue fur), Y14-2 (thirst desire cold drinks), Y25-1 (insomnia, dreamfulness); Y01 (dry stool or constipation), Y04 (spontaneous sweating, dry tongue, fast pulse, fissured tongue), Y07 (fetid mouth odor, swift digestion rapid hungering), Y11 (agitation or short temper), Y19 (tinnitus resemble tide), Y10-2 (red tongue), Y25-2 (flushed face).

- Phlegm-Dampness: Y06 (thick tongue fur, greasy tongue fur), Y11 (sticky or greasy feel in mouth), Y20 (fat tongue, tooth-marked tongue), Y21 (dizziness, head feels as if swathed, distending headache, nausea or vomiting), Y12-1 (slippery pulse), Y14-1 (thirst desire no drinks), Y25-1 (insomnia, dreamfulness); Y03 (dizzy headache), Y16 (urinary incontinence), Y19 (expectoration).

- Yin Deficiency: Y04 (tidal fever or night sweat, spontaneous sweating, dry tongue, fast pulse, fissured tongue), Y07 (vexing heat in chest, edema, fetid mouth odor, muscular twitching, swift digestion rapid hungering), Y19 (sore waist or knees, expectoration, tinnitus resemble cicada),Y31 (blurred vision, dry eyes), Y10-1(pink tongue, hollow headache), Y10-2(red tongue, little tongue fur),Y14-1 (thirst desire no drinks),Y25 (insomnia, dreamfulness, flushed face); Y01 (dry stool or constipation), Y03 (deep-red tongue, thin tongue), Y17 (trembling of limbs), Y21 (dizziness), Y22 (dim complexion, blackish lower eyelid), Y26 (palpitation).

**4.4 Syndrome Quantification**

In the previous section, eight syndrome types were identified as well supported by data. A list of symptom groups was associated with each syndrome type. In the next step, we use those symptom groups to identify patient clusters of that correspond to the syndrome types, and use the population statistics of the patient clusters to quantify the syndrome types. This is done using the Joint Clustering function of Lantern.

**4.4.1 Quantification of Yang Deficiency**

Joint clustering for Yang Deficiency is performed using the model shown in Figure 1. The symptoms are from the symptom groups associated with Yang Deficiency in the previous section. If a group consists of only one symptom variable, then it is directly connected to the joint clustering variable Z at the top. If a group consists of multiple symptom variables, then they are "combined" into a latent feature and the latent feature is connected to Z.

Joint clustering partitions the patients in the VMCI data set into two clusters. Population statistics of the two clusters are as follows:

- Yang Deficiency (0.38): Sore waist or knees (0.77), lassitude of limbs or body (0.69), frequent nocturnal urination (0.68), blackish lower eyelid (0.27), fear of cold or cold limbs (0.44), palpitation (0.45), chest oppression (0.51), thirst desire hot drinks (0.32), spontaneous sweating (0.38).

- Non-Yang Deficiency (0.62): Sore waist or knees (0.21), lassitude of limbs or body (0.25), frequent nocturnal urination (0.28), blackish lower eyelid (0.02), fear of cold or cold limbs (0.12), palpitation (0.13), chest oppression (0.20), thirst desire hot drinks (0.08), spontaneous sweating (0.13).

We see that the symptoms sore waist or knees, lassitude of limbs and body, fear of cold or cold limbs, etc. occur with higher probabilities in the first cluster than in the second cluster. Consequently, the first cluster is interpreted as Yang Deficiency, whereas the second cluster interpreted as non-Yang Deficiency.

According to the results, the prevalence of Yang Deficiency among the patients surveyed is 38%. Among the patients with Yang Deficiency, the symptom sore waist or knees occurs with probability 0.77, lassitude of limbs and body occurs with probability 0.69, fear of cold or cold limbs occurs with probability 0.44, and so on. Among the patients without Yang Deficiency, on the other hand, the corresponding probabilities are 0.21, 0.25, and 0.12 respectively.

Note that, in the quantification given above, the items are ordered in such a way that the symptoms at the front are those whose occurrence probabilities in the two clusters differ the most. Such symptoms are the most important factors to consider when distinguishing between the two clusters. For example, although chest oppression occurs with higher probability than blackish lower eyelid in the Yang Deficiency cluster (0.51 vs 0.27), it is not as important as the latter because it also occurs with higher probability in the non-Yang Deficiency cluster (0.20 vs 0.02).

Moreover, only the nine most important symptoms are used to characterize Yang Deficiency. This is because their cumulative information coverage has reached the cut-off 95% [19] and hence they are sufficient to capture the differences of the two clusters. In contrast, twenty symptoms are listed as relevant to Yang Deficiency in the previous section. Many of them are not important.

### 4.4.2 Quantification of Yin Deficiency

Joint clustering for Yin Deficiency partitions the patients into three clusters. The population statistics of the clusters are as follows. As before, the symptoms are sorted according to their importance in characterizing the differences among the three clusters.

- Yin Deficiency I (0.38): Sore waist or knees (0.75), blurred vision (0.78), dry eyes (0.62), vexing heat in chest (0.08), fetid mouth odor (0.10), tidal fever or night sweat (0.08), insomnia (0.54), tinnitus resemble cicada (0.43), dreamfulness (0.59), spontaneous sweating (0.18).

- Yin Deficiency II (0.08): Sore waist or knees (0.75), blurred vision (0.52), dry eyes (0.37), *vexing heat in chest (0.71), fetid mouth odor (0.71), tidal fever or night sweat (0.71)*, insomnia (0.77), tinnitus resemble cicada (0.43), dreamfulness (0.80), *spontaneous sweating (0.79)*.

- Non-Yin Deficiency (0.54): Sore waist or knees (0.14), blurred vision (0.26), dry eyes (0.16), vexing heat in chest (0.06), fetid mouth odor (0.08), tidal fever or night sweat (0.08), insomnia (0.21), tinnitus resemble cicada (0.10), dreamfulness (0.29), spontaneous sweating (0.19).

The third cluster is interpreted as non-Yin Deficiency, whereas the other two are interpreted as two different subtypes of Yin Deficiency. The main differences between the first two clusters are that the symptoms such as vexing heat in chest, fetid mouth odor, and tidal fever or night sweat occur with much higher probabilities in the second cluster. Therefore, the first cluster is interpreted as Yin Deficiency, and the second cluster Yin Deficiency with Internal Heat.

### 4.4.3 Quantification of Other Syndrome Types

The results of joint clustering for other six syndrome types are listed below. For Blood Deficiency, Blood Stasis, Fire-Heat, and Qi Stagnation, the patients are partitioned into two clusters. For Qi Deficiency and Phlegm-Dampness, the patients are partitioned into three clusters, and two of the clusters are combined because, unlike in the case of Yin Deficiency, the differences between them are not important for treatment.

- Blood Deficiency (0.32): Blurred vision (0.82), dry eyes (0.69), palpitation (0.56), insomnia (0.60), dizziness (0.62), dreamfulness (0.62), numbness (0.53), trembling of limbs (0.20), dry stool or constipation (0.47).

- Non-Blood Deficiency (0.68): Blurred vision (0.32), dry eyes (0.19), palpitation (0.11), insomnia (0.27), dizziness (0.39), dreamfulness (0.36), numbness (0.27), trembling of limbs (0.05), dry stool or constipation (0.23).

- Blood Stasis (0.30): Purple or darkish lips (0.74), dim complexion (0.49), blackish lower eyelid (0.29), numbness (0.58), palpitation (0.42), scaly skin (0.19), tongue with ecchymosis (0.11).

- Non-Blood Stasis (0.70): Purple or darkish lips (0.08), dim complexion (0.10), blackish lower eyelid (0.04), numbness (0.25), palpitation (0.18), scaly skin (0.04), tongue with ecchymosis (0.02).

- Qi Deficiency (0.44): Sore waist or knees (0.70), lack of strength (0.73), lassitude of limbs or body (0.66), short of breath (0.62), chest oppression (0.54), palpitation (0.45), insomnia (0.58), urinary incontinence (0.38), mental fatigue (0.46), dreamfulness (0.61), asthenia of defecation (0.20).

- Non-Qi Deficiency (0.56): Sore waist or knees (0.21), lack of strength (0.29), lassitude of limbs or body (0.23), short of breath (0.20), chest oppression (0.15), palpitation (0.11), insomnia (0.23), urinary incontinence (0.09), mental fatigue (0.18), dreamfulness (0.31), asthenia of defecation (0.05).

- Qi Stagnation (0.35): Chest oppression (0.80), short of breath (0.77), sighing (0.42), hypochondrium distension or pain (0.20), abdominal distension (0.23), dry stool or constipation (0.35).

- Non-Qi Stagnation (0.65): Chest oppression (0.06), short of breath (0.18), sighing (0,12), hypochondrium distension or pain (0.02), abdominal distension (0.07), dry stool or constipation (0.29).

- Fire-Heat (0.31): Dry stool or constipation (0.55), insomnia (0.62), fetid mouth odor (0.30), agitation or short temper (0.79), trembling of limbs (0.24), acid swallow or epigastric upset (0.31), dreamfulness (0.64), spontaneous sweating (0.39), bitter taste in mouth (0.50), aphtha on mouth or tongue

(0.11).

- Non-Fire-Heat (0.69): Dry stool or constipation (0.20), insomnia (0.27), fetid mouth odor (0.06), agitation or short temper (0.48), trembling of limbs (0.03), acid swallow or epigastric upset (0.08), dreamfulness (0.36), spontaneous sweating (0.16), bitter taste in mouth (0.26), aphtha on mouth or tongue (0.01).

- Phlegm-Dampness (0.58): Greasy tongue fur (0.80), slippery pulse (0.60), sticky or greasy feel in mouth (0.29).

- Non-Phlegm-Dampness (0.42): Greasy tongue fur (0.03), slippery pulse (0.27), sticky or greasy feel in mouth (0.05).

### 4.4.4 A Technical Note

We end this section with a subtle technical note concerning joint clustering and we explain the issue using the joint clustering model for Qi Deficiency (Figure 2). Two symptom groups Y26 (chest oppression, palpitation) and Y27 (short of breath) are used in the model among others. In the global model (Figure 3 in [19]), the symptom variable short of breath is grouped, under Y27, with sighing and three other symptom variables. In the joint clustering model, sighing and the other variables are left out. Hence, short of breath might need to be re-grouped with other symptom variables from elsewhere so that the dependence among the variables is properly modeled. In Lantern, there is a search procedure that decides whether re-grouping is necessary and how it should be done. In the joint clustering model for Qi Deficiency, short of breath is automatically merged with Y26 (chest oppression, palpitation), and a new latent variable W is thereby created.

### 4.5 Syndrome Classification

Classification rules are obtained for the eight syndrome types using the Building

Rules function of Lantern, which is explained in [19]. The rule for a syndrome type tells us how to determine whether an individual patient belongs to the syndrome type. Here is the classification rule for Yang Deficiency:

- Yang Deficiency: Sore waist or knees (3.7), lassitude of limbs or body (2.8), frequent nocturnal urination (2.5), blackish lower eyelid (3.8), palpitation (2.5), fear of cold or cold limbs (2.6), chest oppression (2.0), thirst desire hot drinks (2.4), spontaneous sweating (2.0), dim complexion (1.7), undigested food in stool (2.6), muscular twitching (1.4), pale complexion (2.0), diarrhea before dawn (2.1). *Threshold: 9.1; Accuracy: 0.96.*

The rule gives a numerical score for each of the symptoms listed, and a threshold. They are calculated from the symptom occurrence probabilities and syndrome prevalence given in the previous subsection [19]. If the total score of a patient exceeds the threshold, then he is classified into the Yang Deficiency class. Otherwise, he is classified into the non-Yang Deficiency class. The accuracy of the rule is with respect to classification directly based on the joint clustering model [19].

Note that the position of a symptom in the classification rule is determined not only by its score, but also by how frequently it occurs. For example, the score for the symptom undigested food in stool is 2.6, which is the fourth highest. However, the symptom occurs with low probability (<3%) and is hence applicable to only a small fraction of the patients. Consequently, it is placed toward the end of the rule. On the other hand, the symptom frequent nocturnal urination is placed at a much earlier position although its score (2.5) is lower. It occurs with high probability (45%) and hence applies to more patients.

Also note that the last four symptoms in the rule did not appear in the quantification of Yang Deficiency given in the previous section. They occur with low probabilities and hence they are not important factors to consider when describing the main differences between patients with Yang Deficiency and patients without Yang Deficiency. They are included in the rule because, when present, they are important evidence for classifying individual patients.

For Yin Deficiency, there are two classification rules, one for Yin Deficiency versus non-Yin Deficiency, and another for Yin Deficiency I versus Yin Deficiency II (with Internal Heat):

- Yin Deficiency vs Non-Yin Deficiency: Sore waist or knees (4.2), blurred vision (3.0), dry eyes (2.8), tinnitus resemble cicada (2.8), insomnia (2.3), dreamfulness (2.0), expectoration (2.1), blackish lower eyelid (3.2), palpitation (1.9), dizziness (1.6), dry stool or constipation (1.4), vexing heat in chest (1.9), trembling of limbs (2.1), fetid mouth odor (1.7), dim complexion (1.0), tidal fever or night sweat (1.3). *Threshold: 10.6; Accuracy: 0.98.*

- Yin Deficiency II vs Yin Deficiency I: Tidal fever or night sweat (4.9), vexing heat in chest (4.7), fetid mouth odor (4.5), spontaneous sweating (4.1), dry tongue (3.6), edema (3.3), thirst desire no drinks (2.8), fast pulse (3.1), deep-red tongue (3.3), dry stool or constipation (1.9), swift digestion rapid hungering (3.2). *Threshold: 13.9; Accuracy: 0.97.*

The classification rules for the other six syndrome types are as follows:

- Blood Deficiency: Blurred vision (3.4), dry eyes (3.2), palpitation (3.4), insomnia (2.0), dizziness (1.7), dreamfulness (1.6), numbness (1.6), trembling of limbs (2.4), dry stool or constipation (1.5), thin pulse (1.1), muscular twitching (2.1), sallow complexion (1.6), pale lips (1.9), dizzy headache (1.9), pale complexion (2.0). *Threshold: 10.6; Accuracy: 0.98.*

- Blood Stasis: Purple or darkish lips (5.2), dim complexion (3.1), blackish lower eyelid (3.1), numbness (2.0), palpitation (1.7), scaly skin (2.5), tongue with ecchymosis (2.7), darkish tongue (1.0). *Threshold*: 6.4; *Accuracy*: 0.98.

- Qi Deficiency: Sore waist or knees (3.2), lack of strength (2.7), lassitude of limbs or body (2.7), short of breath (2.7), chest oppression (2.8), palpitation (2.8), insomnia (2.2), urinary incontinence (2.7), dreamfulness (1.8), mental fatigue (2.0), asthenia of defecation (2.4), sunken pulse (1.4), dizziness (1.3), spontaneous sweating (1.4), dripping urination (1.7), feeble pulse (2.4), thin pulse (1.2), dizzy headache (2.2). *Threshold: 13.0; Accuracy: 0.96.*

- Qi Stagnation: Chest oppression (5.8), short of breath (3.9), sighing (2.4), hypochondrium distension or pain (3.4), abdominal distension (2.0), dry stool or constipation (0.4). *Threshold*: 6.2; *Accuracy*: 0.97.

- Fire-Heat: Dry stool or constipation (2.2), insomnia (2.1), fetid mouth odor (2.6), agitation or short temper (2.0), trembling of limbs (3.1), acid swallow or epigastric upset (2.4), dreamfulness (1.6), spontaneous sweating (1.9), bitter taste in mouth (1.6), aphtha on mouth or tongue (4.1), dizziness (1.5), dripping urination (1.6), dry tongue (1.7), thirst desire cold drinks (2.5), throbbing headache (2.1). *Threshold*: 9.1; *Accuracy*: 0.94.

- Phlegm-Dampness: Greasy tongue fur (7.1), slippery pulse (2.1), sticky or greasy feel in mouth(2.8), thick tongue fur (1.5), dizzy headache (1.8), tooth-marked tongue(1.0), fat tongue(1.0), urinary incontinence (0.6). *Threshold*: 3.7, *Accuracy*: 0.97.

## 5 Discussions

TCM syndromes are latent variables that cannot be "measured" directly. They can only be "measured" indirectly through symptoms. Based on symptom co-occurrence patterns discovered in data and domain knowledge, we have identified eight groups of symptoms that can be respectively used to "measure" eight syndrome types among patients with VMCI. The patients are partitioned into clusters based on each group of symptoms, and one of the patient clusters are matched up with the syndrome type. The population statistics of the patient cluster are used to quantify the syndrome type (Section 4.4), and to establish a classification rule (Section 4.5).

**5.1 Quality of Results**

There is no ground truth against which to evaluate our results, which was what motivated our work at the first place. However, there are qualitative understandings in TCM theory as to which symptoms are the most important factors to consider when determining whether a patient belongs to a syndrome type. They are called key factors of syndrome differentiation, or *bian zheng yao dian* in Chinese. Our results match those key factors well. They also satisfactorily reflect the characteristics of the VMCI population under study.

**5.1.1 Results on Blood Deficiency**

According to TCM theory, the key factors for determining Blood Deficiency are pale lips, pale or sallow complexion, dizziness, palpitation, insomnia, and thin pulse. All those symptoms are present in our classification rule for Blood Deficiency (Section 4.5). The symptoms with the highest scores are blurred vision and dry eyes. Those reflect the characteristics of the population under study, which consists of patients with VMCI and aged 50 or above. According to TCM theory, seniors usually suffers from kidney deficiency, which can lead to liver Blood Deficiency (liver and kidney are from the same source), and hence poor vision (liver's opening orifice is at eye).

According to our syndrome quantification results (Section 4.4), blurred vision and dry eyes occur with the highest probabilities (0.82 and 0.69) in the Blood Deficiency class. Palpitation and insomnia also occur with high probabilities (0.56 and 0.60). Those suggest that, for the population under study, Blood Deficiency is mainly located at the liver and the heart.

**5.1.2 Results on Blood Stasis**

According to TCM theory, the key factors for determining Blood Stasis are local unpalpable stabbing pain, and color change on the face, the lips and the tongue. In our classification rule for Blood Stasis, concordantly, the symptoms with the highest scores are purple or darkish lips, dim complexion, blackish lower eyelid. The rule also includes tongue with ecchymosis and darkish tongue.

However, our rule does not match TCM theory well regarding local unpalpable stabbing pain. In the VMCI data set, there is only symptom about stabbing pain, i.e., stabbing headache. Stabbing pains at other locations of the body were not included in the survey due to their lack of presences in the TCM literature on *chi dai* or *dai bing*. The symptom stabbing headache occurred with very low frequency (<3%) in the survey and it is negatively, and weakly, correlated with other Blood Stasis symptoms listed above. Its score is -0.4 and is deleted during rule simplification [19].

### 5.1.3 Results on Qi Deficiency

According to TCM theory, the key factors for determining Qi Deficiency are lack of strength, short of breath, and lassitude of limbs and body. In our classification rule for Qi Deficiency, those are among the symptoms with the highest scores. In addition, the four symptoms sore waist or knees, urinary incontinence, chest oppression and palpitation also have high scores. Those reflect the characteristics of the population under study. The patients in the population are seniors and hence are likely to suffer from kidney deficiency, which manifests as sore waist or knees and urinary incontinence. The patients in the population all suffer from vascular MCI (mild cognition impairment). The vascular factors can affect the heart and lead to chest oppression and palpitation.

According to our quantification results, the symptoms chest oppression, palpitation, and sore waist or knees occur with high probabilities in the Qi Deficiency class. Those suggest that, for the population under study, Qi Deficiency is mainly located at the heart and the kidney.

### 5.1.4 Results on Qi Stagnation

According to TCM theory, the key factors for determining Qi Stagnation are gastric distention and fullness, distension pain, and sighing. In our classification rule for blood Qi Stagnation, concordantly, the symptoms with the highest scores are chest oppression, sighing, hypochondrium distension or pain, and abdominal distension.

### 5.1.5 Results on Fire-Heat

According to TCM theory, common manifestations of Fire-Heat include flushed face, thirst, dry stool or constipation, red tongue, yellow tongue fur, and fast pulse. In our classification rule for Fire-Heat, dry stool or constipation and thirst desire cold drinks have high scores. The other top score symptoms are aphtha on mouth or tongue, fetid mouth odor, acid swallow or epigastric upset, insomnia, throbbing headache, and agitation or short temper. In TCM, those symptoms are considered manifestations of excess heat.

However, the symptoms flushed face, red tongue, yellow tongue fur, and fast pulse are not present in our classification rule. They correlate weakly with the aforementioned symptoms in the data set.

### 5.1.6 Results on Phlegm-Dampness

According to TCM theory, the key factors for determining Phlegm-Dampness are the presence of visible phlegm and invisible phlegm, greasy tongue fur, and slippery pulse. Concordantly, greasy tongue fur and slippery pulse have high scores in our classification rule for Phlegm-Dampness. Dizzy headache and dizziness (invisible phlegm) are also present. The important role of dizzy headache and dizziness in the rule is consistent with the TCM claim that the blockage of brain collateral by phlegm is an important pathogenic factor for dementia.

### 5.1.7 Results on Yin Deficiency

According to TCM theory, the manifestations of Yin Deficiency mainly fall into two categories: Those caused by the deprivation of nourishment of viscera and organs due to the lack of Yin fluid, and those caused by the relative excessiveness of Yang due to the failure by Yin to restrain Yang. In our classification rule for Yin Deficiency, the symptoms sore waist or knees, blurred vision, dry eyes and tinnitus resemble cicada are representative symptoms of the first category, while insomnia, dreamfulness, palpitation, expectoration are representative symptoms of the second category.

According to TCM theory, the key characteristics of Yin Deficiency with Internal Heat are red and dry tongue, little tongue fur, and thin and fast pulse. In our rule for differentiating between Yin Deficiency II (with Internal Heat) and Yin Deficiency I, deep-red tongue, dry tongue and fast pulse have high scores. The other top score symptoms are tidal fever night seat, vexing heat in chest, fetid mouth odor, and spontaneous sweating. In TCM theory, those symptoms also suggest Heat caused by relative excessiveness of Yang due to weak Yin.

However, the symptoms little tongue fur and thin pulse are not present in our classification rule. They correlate weakly with the aforementioned symptoms in the data set.

### 5.1.8 Results on Yang Deficiency

Our classification rule for the syndrome type Yang Deficiency matches TCM theory well in the sense that all the symptoms included in the rule are closely related to Yang Deficiency according to TCM theory. Indeed, the symptoms fear of cold or cold limbs, thirst desire hot drink, undigested food in stool, and diarrhea before dawn are the coldness manifestations of Yang Deficiency. The symptoms sore waist or knee, blackish lower eyelids, frequent nocturnal urination, and dim complexion suggest kidney Yang Deficiency. In addition, when Heart Yang fails to warm and push blood, palpitation and chest oppression result in. Weak spleen Yang leads to malnutrition of muscle and meridians, which causes lassitude of limbs and body.

Note that the three symptoms sore waist or knee, blackish lower eyelids, and dim complexion appear both in the rule for Yang Deficiency and the rules for Yin Deficiency. Their total score in the former rule exceeds the threshold, but not in the latter rule. The implication is that the co-occurrence of the three symptoms is sufficient evidence for concluding Yang Deficiency in the VMCI population, while in general one can only conclude kidney deficiency. The reason is that, in the population under study, the three symptoms are always accompanied by some coldness symptoms such as fear of cold or cold limbs, thirst desire hot drink. This is an interesting finding about the population that might be useful for treatment planning.

**5.2 Uses of Results**

Our classification rules are score-based and hence are convenient to use. They can be applied in three scenarios.

First, a major weakness with TCM is the lack of gold standards for syndrome differentiation. In clinic practice, different physicians might draw different conclusions about the syndrome type of the same patient. This affects the efficacy of treatments and undermines patient confidence. The problem can be alleviated substantially if physicians can use the same set of rules as reference. Our classification rules can be used for the purpose. In addition, they can be used as a basis for the development of syndrome differentiation standards for VMCI.

Second, randomized controlled trial (RCT) has been widely used to determine the efficacy of TCM treatments. To conduct RCT on a treatment for a TCM syndrome type, one needs to: (1) identify patients belonging to the syndrome types, and (2) then randomly divide them into treatment and control groups. Our classification rules can be used in the first step for RCT on TCM treatments for VMCI patients.   .

Third, there has been much interest on TCM syndrome essence study, where the objective is to identify biomedical indicators that can be used as gold standards for TCM syndrome differentiation [e.g., 25]. To conduct such research, one needs to: (1) divide patients into those who belong to a syndrome type and those who do not, and (2) compare the biomedical measurements of patients in the two classes. TCM syndrome essence study has so far achieved little success. A key reason is that the first step was not done properly [25]. In our work, the patient clusters are identified based on symptom occurrence patterns hidden in data. If there are indeed biomedical indicators that capture the "essence" of syndrome types, they should correlate well with syndrome occurrence patterns, and hence correlated well with the patient clusters identified by our method. As such, our work might help in finding biomedical indicators with high sensitivity and high specificity for TCM syndrome types.

**5.3 Strengths and Limitations**

In this paper we address the problem of classifying VMCI patients into TCM syndrome types. The results are quantitative. Eight syndrome types are identified. Prevalence and score-based classification rules are obtained for each of the syndrome types.

The results are obtained using the latent tree analysis method. Consequently, they enjoy stronger evidence support than solutions one could get using the alternative approaches, namely the expert standardization and the supervised quantification approaches [19]. In expert standardization, syndrome classification solutions are determined by panels of experts. In supervised quantification, experts are asked to determine the syndrome types of individual patients. In both cases, the results depend heavily on who the experts are. The latent tree analysis method, on the other hand, relies only on symptom data. The results are relatively more objective.

One limitation of our work is that the data were all collected in northern China. It would be interesting to conduct the study in other regions of the country, or to include patients from other parts of the country in the study.

Another limitation is that only dichotomous data were used in our study, and hence severity of symptoms was not taken into consideration. This is mainly for the sake of simplicity. While it is possible and would be interesting to use polytomous data, doing so would substantially complicate the methodology, software, and the results. The literature on latent class analysis has mostly been focused on dichotomous data for the same reason [20, 21]. Since the results are meant to be guidelines for physicians to use as references, rather than rigid standards for physicians to follow meticulously, it is desirable to keep them as simple as possible.

**6 Conclusions**

By analyzing the symptom data collected on 803 VMCI patients aged 50 or above using latent tree models, we have reached the conclusion that there are eight TCM syndrome types among the population: Blood Deficiency, Blood Stasis, Phlegm-Dampness, Fire-Heat, Qi Deficiency, Qi Stagnation, Yang Deficiency, and Yin Deficiency. In addition, two subtypes have been identified for Yin Deficiency, namely Yin Deficiency with and without Internal Heat. The prevalence and the symptom occurrence probabilities for the syndrome types and subtypes have been determined, and corresponding classification rules have been established. The results match TCM theory well and enjoy the strongest evidence support currently possible.

**Competing Interests**

'The authors declare that they have no competing interests, financial or non-financial.

**Authors' contributions**

CF and NLZ were the main forces behind the conception and design of the work, and contributed equally to the paper. BXC,XLJ, RJG and ZGC and YLZ played key roles in data collection and results interpretation, and provided valuable comments on earlier versions of the manuscript. ZRC is part of the team that developed the Lantern software and he implemented the search procedure mentioned in Section 4.4. All authors have read and approved the final manuscript, and agree to be accountable to all aspects of the work.

**Acknowledgements:**

Research on this article was supported by Hong Kong Research Grants Council under grant 16202515, Guangzhou HKUST Fok Ying Tung Research Institute, China Ministry of Science and Technology TCM Special Research Projects Program under grants No.200807011 and No.201007002, Beijing Science and Technology Program under grant No.Z111107056811040, Beijing New Medical Discipline Development Program under grant No.XK100270569,and Beijing University of Chinese Medicine under grant No. 2011-CXTD-23.

**Table 1.** Demographic characteristics of the 803 VMCI patients

|  | Overall （N=803） |
|---|---|
| Age(yr) | 68.02±7.72 |
| Sex | |
|   Male(%) | 388(48.3) |
|   Female(%) | 415(51.7) |
| Education | |
|   Illiterates(%) | 90(11.3) |
|   ≤6(%) | 208(25.9) |
|   6-12(%) | 364(45.3) |
|   >12(%) | 140(17.4) |
| Marital status | |
|   Separated(%) | 1(0.1) |
|   Married(%) | 730(91.0) |
|   Divorced(%) | 1(0.1) |
|   Widowed(%) | 70(8.8) |
| BMI | 24.88±0.11 |
| MoCA | 20.57±3.80 |
| Hypertension | 558(69.5) |
|   Hypertension duration | 13.14±0.48 |
| Diabetes | 223(27.8) |
|   Diabetes duration | 8.12±0.43 |
| Dyslipidemia | 196(24.4) |
|   Dyslipidemia duration | 7.19±0.53 |
| Coronary heart disease | 283(35.2) |
|   Coronary heart disease duration | 10.67±0.56 |
| Cerebrovascular disease | 282(35.1) |
|   Cerebrovascular disease duration | 5.31±0.34 |
| Peripheral vascular disease | 18(2.3) |
|   Peripheral vascular disease duration | 5.76±1.70 |

**Table 2**. TCM symptoms and signs included in questionnaire: Where possible the terms are translated according to the WFCMS standard (http://210.76.97.27/zyy).

**Inspection**

Sallow complexion
Pale complexion
Dim complexion
Flushed face
Pale lips
Purple or darkish lips
Blackish lower eyelid
Scaly skin

**Inquiring**

Distending headache
Stabbing headache
Hollow headache
Throbbing headache
Dizzy headache
Head feels as if swathed
Dizziness
Insomnia
Dreamfulness
Agitation or short temper
Blurred vision
Dry eyes
Tinnitus resemble cicada
Tinnitus resemble tide
Thirst desire hot drinks
Thirst desire cold drinks
Thirst desire no drinks
Dry mouth or throat
Bitter taste in mouth
Bland taste in mouth
Fetid mouth odor
Sticky or greasy feel in mouth
Aphtha on mouth or tongue
Expectoration
Mental fatigue

Short of breath
Lack of strength
Palpitation
Chest oppression
Hypochondrium distension or pain
Sighing
Nausea or vomiting
Vomiting of saliva
Acid swallow or epigastric upset
Anorexia
Swift digestion with rapid hungering
Abdominal distension
Weak loins or sore knees
Lassitude of limbs or body
Numbness
Trembling of limbs
Muscular twitching
Fear of cold or cold limbs
Vexing heat in chest
Edema
Spontaneous sweating
Hemihidrosis
Tidal fever or night sweat
Clear profuse urination
Brownish and scanty urination
Frequent nocturnal urination
Dripping urination
Urinary incontinence
Asthenia of defecation
Dry stool or constipation
Loose stool
Undigested food in stool
Diarrhea before dawn
Loose stool following dry feces

**Tongue**

Pale tongue
Red tongue
Deep red tongue
Darkish tongue
Fissured tongue
Fat tongue
Thin tongue
Dry tongue
Tooth marked tongue
Varicose sublingual veins
Tongue with ecchymosis
White tongue fur
Yellow tongue fur
Little tongue fur
Thick tongue fur
Greasy tongue fur

**Palpation**

Taut pulse
Tense pulse
Slippery pulse
Astringent pulse
Fast pulse
Sunken pulse
Moderate pulse
Thin pulse
Slow pulse
Feeble pulse

**Table 3.** Probabilistic co-occurrence patterns and their TCM interpretations: The numbers following a syndrome type indicate the symptoms in the pattern that the syndrome type explains. For example, Y01 reveals the probabilistic co-occurrence of three symptoms. Qi Deficiency explains the first two symptoms, while Blood Stasis explains symptoms number 2 and 3. Y04 reveals the probabilistic co-occurrence of three other symptoms. Yin Deficiency explains all the symptoms, and hence it is not followed by any numbers.

| Latent Variables | Symptoms and signs | Primary syndrome Elements | Secondary Syndrome Elements |
|---|---|---|---|
| Y01 | Asthenia of defecation, dry stool or constipation, sallow complexion, clear profuse urination | Qi Deficiency (1, 2) | Blood Deficiency (2, 3), Fire-Heat (2), Qi Stagnation(2), Yin Deficiency (2),Yang Deficiency（4） |
| Y03 | Pale lips, pale complexion, dizzy headache, deep-red tongue, thin tongue, loose stool following dry feces, hemihidrosis | Blood Deficiency (1, 2, 3,5,7) | Qi Deficiency (2,3,6), Phlegm-Dampness (3), Yin Deficiency (4,5), Yang Deficiency (2) |
| Y04 | Tidal fever or night sweat, spontaneous sweating, dry tongue, fast pulse, fissured tongue | Yin Deficiency (with internal heat) | Qi Deficiency (2) Heat(2,3,4,5), Yang Deficiency (2) |
| Y06 | Thick tongue fur, greasy tongue fur | Phlegm-Dampness | |
| Y07 | Vexing heat in chest, edema, fetid mouth odor, muscular twitching, swift digestion rapid hungering | Yin Deficiency (with internal heat) | Heat(3,5), Yang Deficiency (4),Blood Deficiency (4) |
| Y08 | Sunken pulse, feeble pulse | Qi Deficiency | |
| Y11 | Sticky or greasy feel in mouth, agitation or short temper | Phlegm-Dampness (1) | Heat (2) |
| Y13 | Varicose sublingual-veins, astringent pulse, tense pulse | Qi Stagnation, Blood Stasis | |
| Y16 | Bitter taste in mouth, urinary incontinence | Heat | Qi Deficiency (2), Phlegm-Dampness (2) |
| Y17 | Acid swallow or epigastric upset, abdominal distension, trembling of limbs | Heat (1,3) | Qi Stagnation (2), Yin Deficiency (3), Blood Deficiency (3) |
| Y18 | Aphtha on month or tongue, throbbing headache | Heat | |
| Y19 | Sore waist or knees, expectoration, tinnitus resemble cicada | Yin Deficiency | Qi Deficiency (1), Yang Deficiency (1), Phlegm-Dampness (2) |
| Y20 | Fat tongue, tongue with ecchymosis, tooth-marked tongue | Qi Deficiency(1,3), Phlegm-Dampness(1,3) | Blood Stasis (2) |
| Y21 | Dizziness, head feels as if swathed, distending headache, nausea or vomiting | Phlegm-Dampness | Heat (1,3,4),Qi Deficiency (1), Blood Deficiency (1), Yin Deficiency(1), Blood Stasis(1) |
| Y22 | Dim complexion, blackish lower eyelid, scaly skin | Blood Stasis | Yang Deficiency (1,2), Yin Deficiency(1,2) |
| Y23 | Purple or darkish lips | Blood Stasis | |
| Y24 | Lassitude of limbs or body, fear of cold or cold limbs, numbness | Yang Deficiency (1, 2) | Qi Deficiency (1),Blood Deficiency (3),Blood Stasis (3) |
| Y26 | Chest oppression, palpitation | Qi Deficiency, Yang Deficiency, | Qi Stagnation(1), Blood Deficiency (2),Blood Stasis (2),Yin Deficiency(2) |
| Y27 | Short of breath, sighing, hypochondrium distension or pain, tinnitus resemble tide, stabbing headache | Qi Stagnation(1,2,3) | Qi Deficiency (1), Fire-Heat(4), Blood Stasis(5) |
| Y29 | Lack of strength, mental fatigue, loose Stool | Qi Deficiency | |
| Y30 | Frequent nocturnal urination, dripping urination | Yang Deficiency(1) | Qi Deficiency (2) |
| Y31 | Blurred vision, dry eyes | Blood Deficiency, Yin Deficiency | |

**Table 4.** Probabilistic mutual-exclusion patterns and their TCM interpretations

| Latent Variables | Symptoms and Signs | Primary Syndrome Elements | Secondary Syndrome Elements |
|---|---|---|---|
| Y02 | Y02-1: Anorexia, undigested food in stool, bland taste in mouth, vomiting of saliva, diarrhea before dawn | Yang Deficiency | Qi Deficiency(1,3) |
|  | Y02-2: Darkish tongue | Blood Stasis |  |
| Y05 | Y05-1: White tongue fur | - |  |
|  | Y05-2: Yellow tongue fur | Heat |  |
| Y09 | Y09-1: Slow pulse, moderate pulse | Qi Deficiency |  |
|  | Y09-2: Taut pulse | Qi Stagnation |  |
| Y10 | Y10-1: Pink tongue, hollow headache | Yin Deficiency |  |
|  | Y10-2: Red tongue, little tongue fur | Yin Deficiency | Heat(1) |
| Y12 | Y12-1: Slippery pulse | Phlegm-Dampness |  |
|  | Y12-2: Thin pulse | Qi Deficiency, Blood Deficiency |  |
| Y14 | Y14-1: Thirst desire no drinks | Yin Deficiency, Phlegm-Dampness |  |
|  | Y14-2: Thirst desire cold drinks | Heat |  |
| Y15 | Y15-1: Thirst desire hot drinks | Yang Deficiency |  |
|  | Y15-2: Not "thirst desire hot drinks" | - |  |
| Y25 | Y25-1: Insomnia, dreamfulness | Yin Deficiency, Fire-Heat, Blood Deficiency, Qi Deficiency, Phlegm-Dampness |  |
|  | Y25-2: Flushed face | Heat, Yin Deficiency |  |

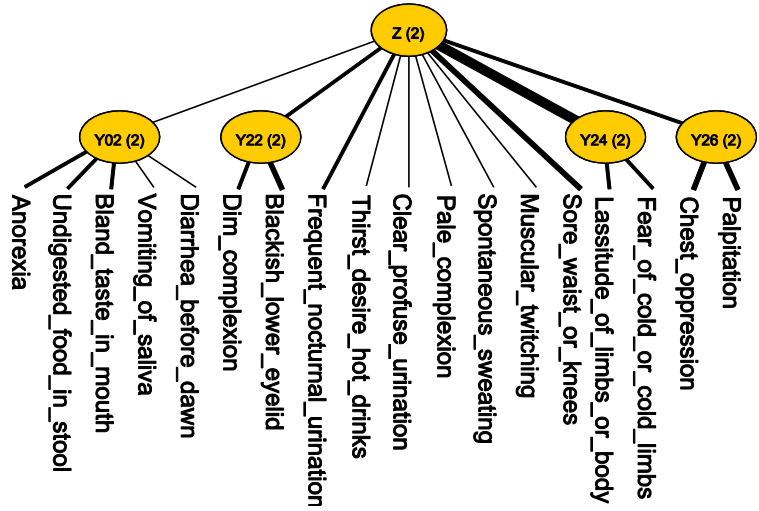

**Figure 1.** Joint clustering model for Yang Deficiency.

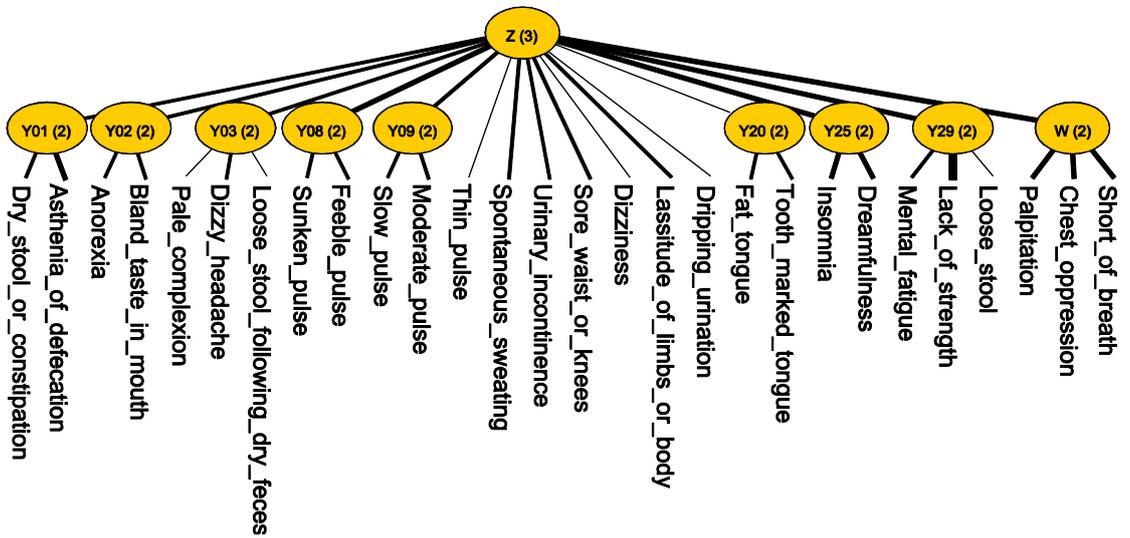

**Figure 2.** Joint clustering model for Qi Deficiency.